# A New Keyphrases Extraction Method Based on Suffix Tree Data Structure for Arabic Documents Clustering

Issam SAHMOUDI, Hanane FROUD and Abdelmonaime LACHKAR

L.S.I.S, E.N.S.A Sidi Mohamed Ben Abdellah University (USMBA) Fez, Morocco

## ABSTRACT

*Document Clustering is a branch of a larger area of scientific study known as data mining .which is an unsupervised classification using to find a structure in a collection of unlabeled data. The useful information in the documents can be accompanied by a large amount of noise words when using Full Text Representation, and therefore will affect negatively the result of the clustering process. So it is with great need to eliminate the noise words and keeping just the useful information in order to enhance the quality of the clustering results. This problem occurs with different degree for any language such as English, European, Hindi, Chinese, and Arabic Language. To overcome this problem, in this paper, we propose a new and efficient Keyphrases extraction method based on the Suffix Tree data structure (KpST), the extracted Keyphrases are then used in the clustering process instead of Full Text Representation. The proposed method for Keyphrases extraction is language independent and therefore it may be applied to any language. In this investigation, we are interested to deal with the Arabic language which is one of the most complex languages. To evaluate our method, we conduct an experimental study on Arabic Documents using the most popular Clustering approach of Hierarchical algorithms: Agglomerative Hierarchical algorithm with seven linkage techniques and a variety of distance functions and similarity measures to perform Arabic Document Clustering task. The obtained results show that our method for extracting Keyphrases increases the quality of the clustering results. We propose also to study the effect of using the stemming for the testing dataset to cluster it with the same documents clustering techniques and similarity/distance measures.*

## KEYWORDS

*Arabic Language, Arabic Text Clustering, Hierarchical Clustering, Suffix Tree Data Structure, Stemming, Keyphrases Extraction.*

## 1. INTRODUCTION

Document clustering is one of the important text mining tasks and now it becomes a natural activity in every organization. It refers to the process of grouping documents with similar contents or topics into clusters using different clustering methods and algorithms.

Traditional documents clustering algorithms use the Full-Text in the documents to generate feature vectors. Such methods often produce unsatisfactory results because there is much noisy information in documents. There is always a need to summarize information into compact form that could be easily absorbed. The challenge is to extract the essence of text documents collections and present it in a compact form named Keyphrases that identifies their topic(s).

Keyphrases are defined as a set of terms in a document that give a brief summary of its content for readers [1]. Keyphrases extraction is a process by which the set of words or phrases that best describe a document is specified. The Keyphrases extraction from free text documents is becoming increasingly important as the uses for such technology expands. It plays a vital role in the task of indexing, summarization, clustering (Turney, 2000), categorization (Hulth and Megyesi 2006) and more recently in improving search results and in ontology learning (Englmeier, Murtaghì et al. 2007) [2].

Work on automatic Keyphrases extraction started rather recently. First essay to approach this goal were based on heuristics by Krulwich et al in 1996[3]. He use heuristics to extract significant phrases from document for learning rather than use standard mathematical techniques. However Keyphrases generated by this approach, failed to map well to author assigned keywords indicating that the applied heuristics were poor ones [3]. Generally, the proposed Keyphrases extraction algorithms may be classified under two categories: supervised approach and unsupervised one. In the next section, these two approaches we will be briefly presented and described.

**The supervised approach:** Turney in 2000 [4] was the first to approach the task of Keyphrases extraction as a supervised learning problem which is called GenEx. He regards Keyphrases extraction as classification task. The GenEx is a Hybrid Genetic Algorithm for Keyphrases extraction which has two components, the Genitor genetic algorithm created by Whitley et al in 1989 [5] and the Extractor which is a Keyphrases extraction algorithm created by Turney in (1997, 2000) [6][4]. The Extractor takes a document as input and produces a list of Keyphrases as output. The Extractor has twelve parameters that determine how it processes the input text. In GenEx, the parameters of the Extractor are tuned by the Genitor genetic algorithm to maximize performance on training data. Genitor is used to tune the Extractor, but Genitor is no longer needed once the training process is complete. Basing on the work of Turney, another algorithm have been created by Witten et al. in 1999 [7] called Keyphrases extraction algorithm (KEA). This latter use a Naïve Bayes as learning model using a set of training documents with known Keyphrases. Then, the model will be used to determine which sentences of an original document are likely to be Keyphrases. Another Keyphrases extraction method has been created by Hulth in 2003 [8 ] as Like GenEx and KEA, this method is different to the above methods, in fact there is no limit on the length of the extracted Keyphrases. In our work, which focuses on the Arabic language, the supervised approach will not useful because of the lack of training data or corpus. Therefore, it will be of great interest to adopt the unsupervised approach for Arabic Keyphrases extraction.

**The unsupervised approach:** Rada [9] propose to use graph-based ranking method which constructs a word graph according to word co-occurrences within the document, and then use random walk techniques to measure word importance. After that, top ranked words are selected as Keyphrases. Unlike other existing Keyphrases extraction systems, the KP-Miner system proposed by El-Beltagy et all [2] may be considered as one of the well known algorithm for Keyphrases extraction especially from Arabic language. The KP-Miner system can be summarised in three logical steps: Candidate Keyphrases selection, Candidate Keyphrases weight calculation and finally Candidate Phrases List Refinement.

From the above brief survey, in our knowledge, up today there is one investigation that has been proposed for Arabic Keyphrases extraction, so we cannot make any comparative study to select the most appropriate method. Therefore, it's with great need to enrich this field of study by proposing others method based on different mathematical models and use different data structures.

In this investigation, in order to enrich this field of study, and therefore to be able to make a comparative study, we propose our novel Keyphrases extraction method based on the Suffix Tree data structure (KpST) especially for Arabic Keyphrases extraction.

The extracted Keyphrases are then used as compact representation of the document in the clustering process instead of the full-Text representation. To illustrate the interest of our proposed Keyphrases extraction method, a series of experiences have been conducted for Arabic documents clustering using Agglomerative Hierarchical algorithm with seven linkage techniques with a variety of distance functions and similarity measures, such as the Euclidean Distance, Cosine Similarity, Jaccard Coefficient, and the Pearson Correlation Coefficient [10][11]. The obtained results show that our proposed Keyphrases extraction method enhances greatly the quality of the clustering process in comparison with the full-Text representation.

The remainder of this paper is organized as follows. The next section discusses our novel approach to extract the Keyphrases using Suffix Tree algorithm for Arabic documents. Section 3 presents the similarity measures and their semantics. Section 4 explains experiment settings, dataset, evaluation approach, results and analysis. Section 5 concludes and discusses our future works.

## 2. PROPOSED ARABIC KEYPHRASES EXTRACTION USING SUFFIX TREE DATA STRUCTURE

Keyphrases are widely used in large document collections. They describe the content of single documents and provide a kind of semantic metadata that is useful for a variety of purposes [12]. Many Keyphrases extractors view the problem as a classification problem and therefore they need training documents (i.e. documents which their Keyphrases are known in advance). Other systems view Keyphrases extraction as a ranking problem. In the latter approach, the words or phrases of a document are ranked based on their importance and phrases with high importance (usually located at the beginning of the list) are recommended as possible Keyphrases for a document.

From the observation of human-assigned Keyphrases, we conclude that good Keyphrases of a document should satisfy the following properties:

- **Understandable.** The Keyphrases are understandable to people. This indicates the extracted Keyphrases should be grammatical.
- **Relevant.** The Keyphrases are semantically relevant with the document theme.
- **Good coverage**. The Keyphrases should cover the whole document well.

In this paper, a novel unsupervised Keyphrases extraction approach based on generalized Suffix Tree construction for Arabic documents is presented and described.

## 2.1 Flowchart of the Proposed Method for Arabic Keyphrases Extraction

Our proposed new method for Arabic Keyphrases Extraction can be presented by the Flowchart (Figure.1) and summarized as follow: In the first phase, the Arabic document is segmented on a set of sentences. Then, each sentence will be cleaned from Arabic stop words, punctuation mark and specials characters like ( /, #, $, ect...). After cleaning phase, a set of sentences is used to generate the Suffix Tree Document Model (STDM) Figure.2 and Figure.3. Each node in STDM is scored and the highest scoring is selected as candidate of Keyphrases. Note that node will be considered as keyphrases if only their label length between one and three.

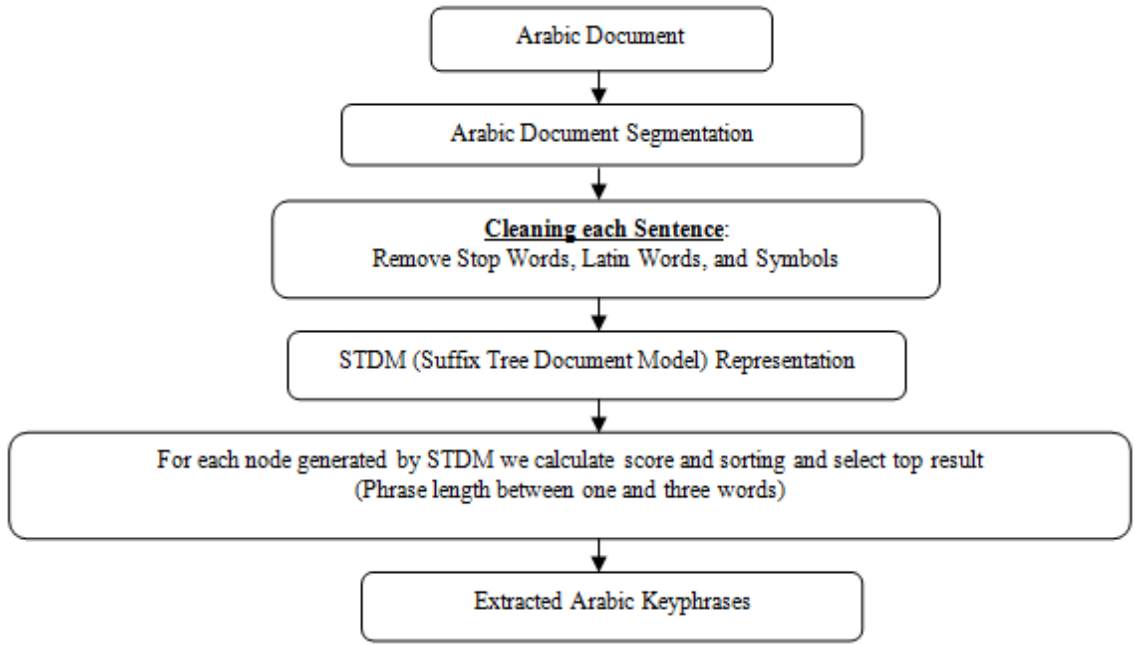

Fig.1. Flowchart of the Proposed Method for Arabic Keyphrases Extraction

## 2.1.1 Suffix Tree Data Structure

A Suffix Tree is a data structure that allows efficient string matching and querying. It have been studied and used extensively, and have been applied to fundamental string problems such as finding the longest repeated substring, strings comparisons , and text compression[13]. The Suffix Tree commonly deals with strings as sequences of characters, or with documents as sequences of words. A Suffix Tree of a string is simply a compact tree of all the suffixes of that string. The Suffix Tree has been used firstly by Zamir et al. [14] as clustering algorithm named Suffix Tree Clustering (STC). It's linear time clustering algorithm that is based on identifying the shared phrases that are common to some Document's in order to group them in one cluster. A phrase in our context is an ordered sequence of one or more words. Suffix Tree has two logical steps: Document's "Cleaning", and Identifying Suffix Tree Document Model.

### a. Document's "Cleaning"

In this step, each document is processed for Arabic stop-words removal such as (e.g., والذي وان فانه وهذا فكان ستكون...): Stop-word means high frequency and low discrimination and should be filtered out in the IR system. They are functional, general, and common words of the language that usually do not contribute to the semantics of the documents and have no read added value. Many Information retrieval systems (IRS) attempt to exclude stop-words from the list of features to reduce feature space, and to increase their performance. In addition, in our case, to deal especially with Arabic snippets, we propose also in this step to remove Latin words and specials characters such as (e.g. $, #,...)[15].

**b. Suffix Tree Document Model**

The Suffix Tree treats documents as a set of phrases (sentences) not just as a set of words. The sentence has a specific semantic meaning (words in the sentence are ordered). Suffix Tree Document Model (STDM) considers a document $d = w_1 w_2 \ldots w_m$ as a string consisting of words $w_i$, not characters (i = 1; 2;…; m). A revised definition of suffix tree can be presented as follow: A Generalized Suffix Tree for a set S of n strings, each of length $m_n$, is a rooted directed tree with exactly $\sum m_n$ leaves marked by a two number index (k,$l$) where k ranges from 1 to n and $l$ ranges from 1 to $m_k$. Each internal node, other than the root, has at least two children and each edge is labelled with a nonempty substring of words of a string in S. No two edges out of a node can have edge labels beginning with the same word. For any leaf (i,j), the concatenation of the edge labels on the path from the root to leaf(i, j) exactly spells out the suffix of $S_i$ that starts at position j , that's it spells out $S_i[ j…m_i ][14]$.The Figure.2 shows an example of the generated Suffix Tree of a set of three Arabic strings or three Documents–Document1: "القط يأكل الجبن" , Document2: "الفارياكل الجبن ايضا", Document3: "القط ياكل الفارايضا", respectively the Figure.3 shows the same example in English Language (Document1: "**cat ate cheese**", Document2: "**mouse ate cheese too** " and Document3: "**cat ate mouse too**") .

**2.1.2 Score Calculation:**

A Suffix Tree of document d is a compact tree containing all suffixes of document d, this tree is presented by a set of nodes and leaves and labels. The label of a node in the tree is defined as the concatenation, in order, of the sub-strings labeling the edges of the path from the root to that node. Each node should have a score. The node can be ranked according to its score.
The score is relevant to:

    a. The length of the label node words.

    b. The number of the occurrence of word in document Term Frequency

Each node of the suffix tree is scored as:

$$S(B) = |B|*F(|P|) * \sum TFIDF (w_i) \qquad (1)$$

$$F(P) = \begin{cases} |P|, \text{ if } 3 \geq |P| \geq 1 \\ \\ 0, \text{ Otherwise} \end{cases} \qquad (2)$$

Where |B| is the number of documents shared the phrase P, and |P| is the number of words making up the phrase P, $w_i$ represents the words in P and TFIDF represent the Term Frequency Inverse Document Frequency for each word $w_i$ in P.

Fig.2. Suffix Tree for Arabic example (A,B,C)

Fig.3. Suffix Tree using English example (A,B,C)

- Example of calculating score of phrase A = "cat ate" using STDM figurate in Figure 2:

- Number of documents share A = 2

- Length of A= 2

- ∑TFIDF (cat ate) = (1/3+1/4)*log(3/2+1))+((1/3+1/4+1/4)log(2))= 0.2508

$$S(A) = 1.0032$$

## 3. SIMILARITY MEASURES

In this section we present the five similarity measures that were tested in [10] and our works [11][16], and we include these five measures in our work to effect the Arabic text document clustering.

### 3.1. Euclidean Distance

Euclidean distance is widely used in clustering problems, including clustering text. It satisfies all the above four conditions and therefore is a true metric. It is also the default distance measure used with the K-means algorithm. Measuring distance between text documents, given two documents $d_a$ and $d_b$ represented by their term vectors $\vec{t_a}$ and $\vec{t_b}$ respectively, the Euclidean distance of the two documents is defined as:

$$D_E(\vec{t_a}, \vec{t_b}) = (\sum_{t=1}^{m} |w_{t,a} - w_{t,b}|^2)^{1/2}, \qquad (3)$$

where the term set is $T = \{t_1, ..., t_m\}$. As mentioned previously, we use the $tfidf$ value as term weights, that is $w_{t,a} = tfidf(d_a, t)$.

### 3.2. Cosine Similarity

Cosine similarity is one of the most popular similarity measure applied to text documents, such as in numerous information retrieval applications [17] and clustering too [18].

Given two documents $\vec{t_a}$ and $\vec{t_b}$, their cosine similarity is:

$$SIM_C(\vec{t_a}, \vec{t_b}) = \frac{\vec{t_a} \cdot \vec{t_b}}{|\vec{t_a}| \times |\vec{t_b}|}. \qquad (4)$$

where $\vec{t_a}$ and $\vec{t_b}$ are m-dimensional vectors over the term set $T = \{t_1, ..., t_m\}$. Each dimension represents a term with its weight in the document, which is non-negative. As a result, the cosine similarity is non-negative and bounded between $[0,1]$. An important property of the cosine similarity is its independence of document length. For example, combining two identical copies of a document d to get a new pseudo document $d_0$, the cosine similarity between d and $d_0$ is 1, which means that these two documents are regarded to be identical.

### 3.3. Jaccard Coefficient

The Jaccard coefficient, which is sometimes referred to as the Tanimoto coefficient, measures similarity as the intersection divided by the union of the objects. For text document, the Jaccard coefficient compares the sum weight of shared terms to the sum weight of terms that are present in either of the two documents but are not the shared terms. The formal definition is:

$$SIM_J(\vec{t_a}, \vec{t_b}) = \frac{\vec{t_a} \cdot \vec{t_b}}{|\vec{t_a}|^2 + |\vec{t_b}|^2 - \vec{t_a} \cdot \vec{t_b}} \qquad (5)$$

The Jaccard coefficient is a similarity measure and ranges between 0 and 1. It is 1 when the $\vec{t_a} = \vec{t_b}$ and 0 when $\vec{t_a}$ and $\vec{t_b}$ are disjoint. The corresponding distance measure is $D_J = 1 - SIM_J$ and we will use $D_J$ instead in subsequent experiments.

### 3.4. Pearson Correlation Coefficient

Pearson's correlation coefficient is another measure of the extent to which two vectors are related. There are different forms of the Pearson correlation coefficient formula. Given the term set $T = \{t_1, ..., t_m\}$, a commonly used form is:

$$SIM_P(\vec{t_a}, \vec{t_b}) = \frac{m \sum_{t=1}^{m} w_{t,a} \times w_{t,b} - TF_a \times TF_b}{\sqrt{\left[ m \sum_{t=1}^{m} w_{t,a}^2 - TF_a^2 \right] \left[ m \sum_{t=1}^{m} w_{t,b}^2 - TF_b^2 \right]}} \qquad (6)$$

where $TF_a = \sum_{t=1}^{m} w_{t,a}$ and $TF_b = \sum_{t=1}^{m} w_{t,b}$

This is also a similarity measure. However, unlike the other measures, it ranges from -1 to +1 and it is 1 when $\vec{t_a} = \vec{t_b}$. In subsequent experiments we use the corresponding distance measure, which is $D_P = 1 - SIM_P$ when $SIM_P \geq 0$ and $D_P = |SIM_P|$ when $SIM_P \prec 0$.

### 3.5. Averaged Kullback-Leibler Divergence

In information theory based clustering, a document is considered as a probability distribution of terms. The similarity of two documents is measured as the distance between the two corresponding probability distributions. The Kullback-Leibler divergence (KL divergence), also called the relative entropy, is a widely applied measure for evaluating the differences between two probability distributions. Given two distributions P and Q, the KL divergence from distribution P to distribution Q is defined as:

$$D_{KL}(P \| Q) = P \log(\frac{P}{Q}) \qquad (7)$$

In the document scenario, the divergence between two distributions of words is:

$$D_{KL}(\vec{t_a} \| \vec{t_b}) = \sum_{t=1}^{m} w_{t,a} \times \log(\frac{w_{t,a}}{w_{t,b}}) \qquad (8)$$

However, unlike the previous measures, the KL divergence is not symmetric, i.e. $D_{KL}(P \| Q) \neq D_{KL}(Q \| P)$. Therefore it is not a true metric. As a result, we use the averaged KL divergence instead, which is defined as:

$$D_{AvgKL}(P \| Q) = \pi_1 D_{KL}(P \| M) + \pi_2 D_{KL}(Q \| M), \qquad (9)$$

where $\pi_1 = \frac{P}{P+Q}, \pi_2 = \frac{Q}{P+Q}$ and $M = \pi_1 P + \pi_2 Q$ For documents, the averaged KL divergence can be computed with the following formula:

$$D_{AvgKL}(\vec{t_a} \| \vec{t_b}) = \sum_{t=1}^{m} (\pi_1 \times D(w_{t,a} \| w_t) + \pi_2 \times D(w_{t,b} \| w_t)); \qquad (10)$$

where $\pi_1 = \frac{w_{t,a}}{w_{t,a} + w_{t,b}}, \pi_2 = \frac{w_{t,b}}{w_{t,a} + w_{t,b}},$ and $w_t = \pi_1 \times w_{t,a} + \pi_2 \times w_{t,b}$.

The average weighting between two vectors ensures symmetry, that is, the divergence from document i to document j is the same as the divergence from document j to document i. The averaged KL divergence has recently been applied to clustering text documents, such as in the family of the Information Bottleneck clustering algorithms [19], to good effect.

# 4. EXPERIMENTS AND RESULTS

The evaluation of our new method for Arabic Keyphrases extraction can be presented by the Flowchart (Figure.4) and summarized as follow: From Heterogeneous dataset of Arabic documents we build two types of dataset the Full-Text Dataset and Keyphrases Dataset. So therefore, the second type will be constructed by applying our Keyphrases extraction on each document and it will be composed by the documents results. The two types of datatsets will be clustered using the most popular Clustering approach of Hierarchical algorithms with seven linkage techniques and a variety of distance functions and similarity measures to perform Arabic Document Clustering task. The Figure 5 presents the same experiments steps as Figure.4 by using the stemming process before Keyphrases extraction in order to study it impact on the clustering results. Note that, the used similarity measures do not directly fit into the algorithms, because smaller values indicate dissimilarity [11]. The Euclidean distance and the Averaged KL Divergence are distance measures, while the Cosine Similarity, Jaccard coefficient and Pearson coefficient are similarity measures. We apply a simple transformation to convert the similarity measure to distance values. Because both Cosine Similarity and Jaccard coefficient are bounded in $[0,1]$ and monotonic, we take $D = 1 - SIM$ as the corresponding distance value. For Pearson coefficient, which ranges from $-1$ to $+1$, we take $D = 1 - SIM$ when $SIM \geq 0$ and $D = |SIM|$ when $SIM \prec 0$.

For the testing dataset, we experimented with different similarity measures for two times: without stemming, and with stemming using the Morphological Analyzer from Khoja and Garside [20] , in two case: in the first one, we apply the proposed method above to extract Keyphrases for the all documents in dataset and then cluster them. In the second case, we cluster the original documents. Moreover, each experiment was run for many times and the results are the averaged value over many runs. Each run has different initial seed sets; in the total we had 105 experiments for Agglomerative Hierarchical algorithm using 7 techniques for merging the clusters described below in the next section.

## 4.1. Agglomerative Hierarchical Techniques

Agglomerative algorithms are usually classified according to the inter-cluster similarity measure they use. The most popular of these are [17][18]:

- Linkage: minimum distance criterion

$$d_{A \to B} = \min_{\substack{\forall i \ \varepsilon A \\ \forall j \ \varepsilon B}} (d_{ij}) \tag{11}$$

- Complete Linkage : maximum distance criterion

$$d_{A \to B} = \max_{\substack{\forall i \ \varepsilon A \\ \forall j \ \varepsilon B}} (d_{ij}) \tag{12}$$

- Average Group : average distance criterion

$$d_{A \to B} = \underset{\substack{\forall i \, \in A \\ \forall j \, \in B}}{average}(d_{ij}) \qquad (13)$$

- Centroid Distance Criterion :

$$d_{A \to B} = \| C_A - C_B \| = \frac{1}{n_i n_j} \sum_{\substack{\forall i \, \in A \\ \forall j \, \in B}} (d_{ij}) \qquad (14)$$

- Ward : minimize variance of the merge cluster.

Jain and Dubes (1988) showed general formula that first proposed by Lance and William (1967) to include most of the most commonly referenced hierarchical clustering called SAHN (Sequential, Agglomerative, Hierarchical and nonoverlapping) clustering method. Distance between existing cluster k with nk objects and newly formed cluster (r,s) with nr and ns objects is given as:

$$d_{k \to (r,s)} = \alpha_r d_{k \to r} + \alpha_s d_{k \to s} + \beta d_{r \to s} + \gamma \left| d_{k \to r} - d_{k \to s} \right| \qquad (15)$$

The values of the parameters are given in the in Table 1

Table I : The Values of the Parameters of the General Formula of Hierarchical Clustering SAHN

| Clustering Method | $\alpha_r$ | $\alpha_s$ | $\beta$ | $\gamma$ |
|---|---|---|---|---|
| Single Link | ½ | 1/2 | 0 | -1/2 |
| Complete Link | ½ | 1/2 | 0 | 1/2 |
| Unweighted Pair Group Method Average (UPGMA) | $\dfrac{n_r}{n_r + n_s}$ | $\dfrac{n_s}{n_r + n_s}$ | 0 | 0 |
| Weighted Pair Group Method Average (WPGMA) | ½ | 1/2 | 0 | 0 |
| Unweighted Pair Group Method Centroid (UPGMC) | $\dfrac{n_r}{n_r + n_s}$ | $\dfrac{n_s}{n_r + n_s}$ | $\dfrac{-n_r n_s}{(n_r + n_s)^2}$ | 0 |
| Weighted Pair Group Method Centroid (WPGMC) | ½ | 1/2 | -1/4 | 0 |
| Ward's Method | $\dfrac{n_r + n_k}{n_r + n_s + n_k}$ | $\dfrac{n_s + n_k}{n_r + n_s + n_k}$ | $\dfrac{-n_k}{n_r + n_s + n_k}$ | 0 |

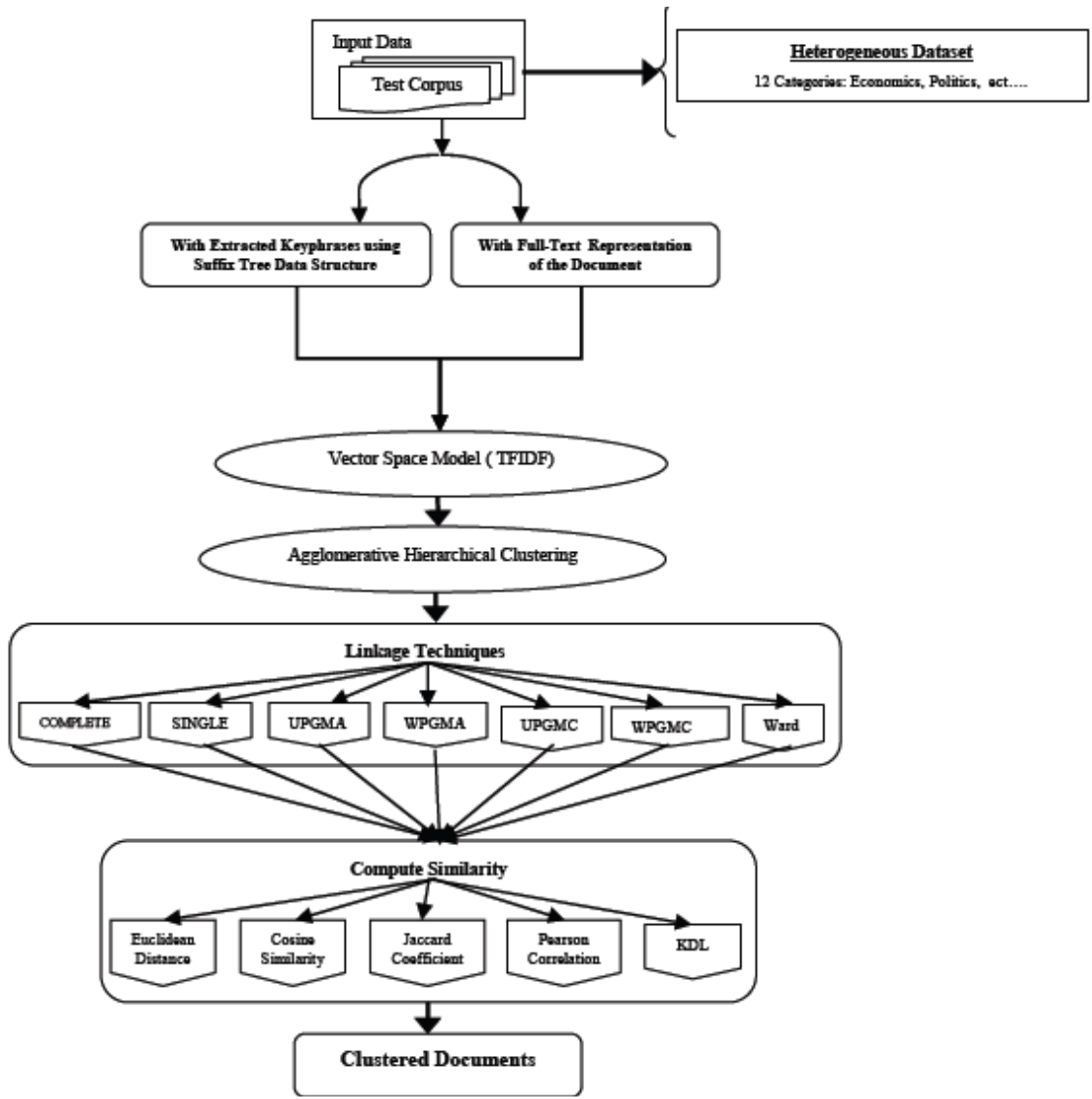

Fig.5.Description of Our Experiments Stemmed Documents

## 4.2. Dataset

The testing dataset (Corpus of Contemporary Arabic (CCA)) is composed of 12 several categories, each latter contains documents from websites and from radio Qatar. A summary of the testing dataset is shown in Table 2. To illustrate the benefits of our proposed approach, we extracted the appropriate keyphrases from the Arabic documents in our testing dataset using this approach for two times: with and without stemming, and we ranked terms by their weighting schemes $Tfidf$ and use them in our experiments.

Table II. Number of Texts and Number of Terms in Each Category of the Testing Dataset

| Text Categories | Number of Texts | Number of Terms |
|---|---|---|
| Economics | 29 | 67 478 |
| Education | 10 | 25 574 |
| Health and Medicine | 32 | 40 480 |
| Interviews | 24 | 58 408 |
| Politics | 9 | 46 291 |
| Recipes | 9 | 4 973 |
| Religion | 19 | 111 199 |
| Science | 45 | 104 795 |
| Sociology | 30 | 85 688 |
| Spoken | 7 | 5 605 |
| Sports | 3 | 8 290 |
| Tourist and Travel | 61 | 46 093 |

## 4.3. Results and Discussion

The quality of the clustering result was evaluated using two evaluation measures: purity and entropy, which are widely used to evaluate the performance of unsupervised learning algorithms [21], [22]. On the one hand, the higher the purity value (P (cluster) =1), the better the quality of the cluster is. On the other hand, the smaller the entropy value (E (cluster) =0), the better the quality of the cluster is.

Tables 3-10 show the average purity and entropy results for each similarity/distance measure with the Morphological Analyzer from Khoja and Garside [20], and without stemming with document clustering algorithms cited above.

The goal of these experiments is to decide what are the best and appropriate techniques to use for producing consistent clusters for Arabic Documents, with and without using the extracted Keyphrases produced by our novel approach then, the obtained results will be compared with our previous works[23][24].

### 4.3.1 Results without Stemming

The overall entropy and purity values, for our experiments without stemming using the Agglomerative Hierarchical algorithm with 7 schemes for merging the clusters, are shown in the tables 3, 4, 5 and 6.

Tables 3 and 5 summarize the obtained entropy scores in the all experiments, we remarked that the scores shown in the first one are generally worst than those in the second gotten using the extracted Keyphrases, but for those two tables the Agglomerative Hierarchical algorithm performs good using the COMPLETE, UPGMA, WPGMA schemes, and Ward function with the Cosine Similarity, the Jaccard measures and Pearson Correlation. The same behaviour can be concluded from purity scores tables.

### 4.3.2 Results with Stemming

Tables 7, 8, 9 and 10 present the results of using Agglomerative Hierarchical algorithms as document clustering methods with stemming using the Morphological Analyzer from Khoja and Garside [20] for the all documents in dataset.

A closer look to the Tables 7 and 9, lead as to observe the lower entropy scores obtained with the Cosine Similarity, the Jaccard and the Pearson Correlation measures using the COMPLETE, UPGMA, WPGMA schemes, and Ward function as linkage techniques, and the all obtained results shown in the two tables proves that the use of the extracted Keyphrases instead of the full-text representation of documents are slightly better in generating more coherent clusters.

In Tables 8 and 10, the use of Khoja's Stemmer for the all documents clustered by the Agglomerative Hierarchical algorithm  produce good  purity scores for the Cosine Similarity, the Jaccard and the Pearson Correlation measures using the COMPLETE, UPGMA, WPGMA schemes, and Ward function as linkage techniques, when we used the extracted Keyphrases.

The above obtained results (shown in the different Tables) lead us to conclude that:

- For the Agglomerative Hierarchical algorithm, the use of the COMPLETE, UPGMA [16], WPGMA schemes, and Ward function as linkage techniques yield good results.
- Cosine Similarity, Jaccard and Pearson Correlation measures perform better relatively to the other measures for two times: without stemming, and with stemming using the Morphological Analyzer from Khoja and Garside [20].
- The tested documents clustering technique perform well without using the stemming.
- The obtained overall entropy values shown in the different tables proves that the extracted Keyphrases can make their topics salient and improve the clustering performance [23] for two times: with and without stemming.

### 4.3.3 Discussion

The above results shows that, the use of stemming affects negatively the clustering in the two cases (with and without extracted Keyphrases), this is mainly due to the ambiguity created when we applied the stemming (for example, we can obtain two roots that made of the same letters but semantically different); this observation broadly agrees with our previous works [11][16]. Overall, the use of the extracted Keyphrases instead of the full-text representation of documents is the best performing when clustered our Arabic documents that we investigated. There is also another issue that must be mentioned, our experiments show the improvements of the clustering quality and time. In the following, we make a few brief comments on the behaviour of the all tested linkage techniques:

The COMPLETE linkage technique is non-local, the entire structure of the clustering can influence merge decisions. This results in a preference for compact clusters with small diameters, but also causes sensitivity to outliers.

The Ward function allows us to minimize variance of the merge cluster; the variance is a measure of how far a set of data is spread out. So the Ward function is a non-local linkage technique.

With the two techniques described above, a single document far from the center can increase diameters of candidate merge clusters dramatically and completely change the final clustering. That why these techniques produce good results than UPGMA [25], WPGMA schemes and better than the other all tested linkage techniques; because this merge criterion give us local information. We pay attention solely to the area where the two clusters come closest to each other. Other, more distant parts of the cluster and the clusters overall structure are not taken into account.

TABLE III. ENTROPY RESULTS WITHOUT STEMMING USING AGGLOMERATIVE HIERARCHICAL ALGORITHM WITH FULL-TEXT REPRESENTATION

|  | Euclidean | Cosine | Jaccard | Pearson | KLD |
|---|---|---|---|---|---|
| COMPLETE | 0.872 | 0.219 | 0.136 | 0.109 | 0.879 |
| SINGLE | 0.881 | 0.881 | 0.877 | 0.877 | 0.876 |
| UPGMA | 0.872 | 0.329 | 0.064 | 0.116 | 0.879 |
| WPGMA | 0.881 | 0.255 | 0.131 | 0.367 | 0.879 |
| UPGMC | 0.872 | 0.869 | 0.885 | 0.877 | 0.879 |
| WPGMC | 0.881 | 0.885 | 0.884 | 0.884 | 0.879 |
| Ward | 0.699 | 0.100 | 0.091 | 0.073 | 0.707 |

TABLE IV. PURITY RESULTS WITHOUT STEMMING USING AGGLOMERATIVE HIERARCHICAL ALGORITHM WITH FULL-TEXT REPRESENTATION

|  | Euclidean | Cosine | Jaccard | Pearson | KLD |
|---|---|---|---|---|---|
| COMPLETE | 0.878 | 0.556 | 0.558 | 0.535 | 0.933 |
| SINGLE | 0.933 | 0.933 | 0.933 | 0.933 | 0.933 |
| UPGMA | 0.878 | 0.725 | 0.611 | 0.750 | 0.933 |
| WPGMA | 0.933 | 0.749 | 0.748 | 0.823 | 0.933 |
| UPGMC | 0.878 | 0.913 | 0.933 | 0.892 | 0.933 |
| WPGMC | 0.933 | 0.933 | 0.933 | 0.933 | 0.933 |
| Ward | 0.753 | 0.458 | 0.537 | 0.457 | 0.818 |

TABLE V. ENTROPY RESULTS WITHOUT STEMMING USING AGGLOMERATIVE HIERARCHICAL ALGORITHM WITH EXTARCTED KEYPHRASES

|  | Euclidean | Cosine | Jaccard | Pearson | KLD |
|---|---|---|---|---|---|
| COMPLETE | 0.829 | 0.096 | 0.101 | 0.231 | 0.853 |
| SINGLE | 0.854 | 0.857 | 0.859 | 0.854 | 0.853 |
| UPGMA | 0.854 | 0.439 | 0.535 | 0.134 | 0.853 |
| WPGMA | 0.854 | 0.369 | 0.518 | 0.313 | 0.853 |
| UPGMC | 0.854 | 0.679 | 0.337 | 0.851 | 0.853 |
| WPGMC | 0.854 | 0.864 | 0.866 | 0.856 | 0.853 |
| Ward | 0.119 | 0.104 | 0.105 | 0.099 | 0.423 |

TABLE VI. PURITY RESULTS WITHOUT STEMMING USING AGGLOMERATIVE HIERARCHICAL ALGORITHM WITH EXTARCTED KEYPHRASES

|  | Euclidean | Cosine | Jaccard | Pearson | KLD |
|---|---|---|---|---|---|
| COMPLETE | 0.853 | 0.563 | 0.547 | 0.622 | 0.935 |
| SINGLE | 0.935 | 0.935 | 0.935 | 0.935 | 0.935 |
| UPGMA | 0.935 | 0.634 | 0.715 | 0.673 | 0.935 |
| WPGMA | 0.935 | 0.715 | 0.742 | 0.739 | 0.935 |
| UPGMC | 0.935 | 0.831 | 0.829 | 0.894 | 0.935 |
| WPGMC | 0.935 | 0.935 | 0.934 | 0.935 | 0.935 |
| Ward | 0.698 | 0.415 | 0.446 | 0.433 | 0.776 |

TABLE VII. ENTROPY RESULTS WITH KHOJA'S STEMMER USING AGGLOMERATIVE HIERARCHICAL ALGORITHM WITH FULL-TEXT REPRESENTATION

|  |  | Euclidean | Cosine | Jaccard | Pearson | KLD |
|---|---|---|---|---|---|---|
| Khoja's | COMPLETE | 0.878 | 0.172 | 0.165 | 0.213 | 0.875 |
|  | SINGLE | 0.882 | 0.887 | 0.888 | 0.889 | 0.877 |
|  | UPGMA | 0.881 | 0.075 | 0.590 | 0.250 | 0.878 |
|  | WPGMA | 0.882 | 0.319 | 0.446 | 0.156 | 0.875 |
|  | UPGMC | 0.882 | 0.870 | 0.859 | 0.794 | 0.878 |
|  | WPGMC | 0.882 | 0.887 | 0.887 | 0.881 | 0.878 |
|  | Ward | 0.631 | 0.127 | 0.117 | 0.109 | 0.649 |

TABLE VIII. PURITY RESULTS WITH KHOJA'S STEMMER USING AGGLOMERATIVE HIERARCHICAL ALGORITHM WITH FULL-TEXT REPRESENTATION

|  |  | Euclidean | Cosine | Jaccard | Pearson | KLD |
|---|---|---|---|---|---|---|
| Khoja's | COMPLETE | 0.892 | 0.481 | 0.529 | 0.417 | 0.892 |
|  | SINGLE | 0.933 | 0.933 | 0.933 | 0.932 | 0.933 |
|  | UPGMA | 0.891 | 0.578 | 0.751 | 0.559 | 0.933 |
|  | WPGMA | 0.933 | 0.793 | 0.695 | 0.601 | 0.892 |
|  | UPGMC | 0.933 | 0.919 | 0.888 | 0.878 | 0.933 |
|  | WPGMC | 0.933 | 0.933 | 0.932 | 0.891 | 0.933 |
|  | Ward | 0.765 | 0.445 | 0.465 | 0.392 | 0.819 |

TABLE IX. ENTROPY RESULTS WITH KHOJA'S STEMMER USING AGGLOMERATIVE HIERARCHICAL ALGORITHM WITH EXTARCTED KEYPHRASES

|  |  | Euclidean | Cosine | Jaccard | Pearson | KLD |
|---|---|---|---|---|---|---|
| Khoja's | COMPLETE | 0.852 | 0.325 | 0.098 | 0.222 | 0.848 |
|  | SINGLE | 0.854 | 0.861 | 0.851 | 0.863 | 0.853 |
|  | UPGMA | 0.856 | 0.485 | 0.517 | 0.431 | 0.848 |
|  | WPGMA | 0.844 | 0.592 | 0.392 | 0.245 | 0.853 |
|  | UPGMC | 0.856 | 0.860 | 0.841 | 0.864 | 0.848 |
|  | WPGMC | 0.856 | 0.862 | 0.857 | 0.864 | 0.853 |
|  | Ward | 0.539 | 0.139 | 0.086 | 0.150 | 0.556 |

TABLE X. PURITY RESULTS WITH KHOJA'S STEMMER USING AGGLOMERATIVE HIERARCHICAL ALGORITHM WITH EXTARCTED KEYPHRASES

|  |  | Euclidean | Cosine | Jaccard | Pearson | KLD |
|---|---|---|---|---|---|---|
| Khoja's | COMPLETE | 0.907 | 0.561 | 0.494 | 0.525 | 0.936 |
|  | SINGLE | 0.935 | 0.935 | 0.879 | 0.935 | 0.935 |
|  | UPGMA | 0.935 | 0.631 | 0.659 | 0.684 | 0.936 |
|  | WPGMA | 0.873 | 0.778 | 0.689 | 0.754 | 0.935 |
|  | UPGMC | 0.935 | 0.879 | 0.922 | 0.935 | 0.936 |
|  | WPGMC | 0.935 | 0.935 | 0.935 | 0.935 | 0.935 |
|  | Ward | 0.761 | 0.365 | 0.512 | 0.407 | 0.781 |

# 5. CONCLUSIONS

In this paper, we have proposed to illustrate the benefits of using Keyphrases extraction, by comparing the clustering results based on extracted Keyphrases with the full-text baseline on the Arabic Documents Clustering for the most popular approach of Hierarchical algorithms: Agglomerative Hierarchical algorithm using seven linkage techniques with five similarity/distance measures with and without stemming for Arabic Documents.

Our results indicate that the Cosine Similarity, the Jaccard and the Pearson Correlation measures have comparable effectiveness and performs better relatively to the other measures for all documents clustering algorithms cited above for finding more coherent clusters in case we didn't use the stemming for the testing dataset.

Furthermore, our experiments with different linkage techniques for yield us to conclude that COMPLETE, UPGMA, WPGMA and Ward produce efficient results than other linkage techniques. A closer look to these results, show that Ward technique is the best in all cases (with and without using the stemming), although the two other techniques are often not much worse.

Instead of using full-text as the representation for Arabic documents clustering, we use our novel approach based on Suffix Tree algorithm as Keyphrases extraction technique to eliminate the noise on the documents and select the most salient sentences to represent the original documents. Furthermore, Keyphrases extraction can help us to overcome the varying length problem of the diverse documents.

In our experiments using Keyphrases extraction, we remark that again the Cosine Similarity, the Jaccard and the Pearson Correlation measures have comparable effectiveness to produce more coherent clusters than the Euclidean Distance and averaged KL Divergence, in the two times: with and without stemming, on the other hand, the good results are detected when using COMPLETE, UPGMA, WPGMA and Ward as linkage techniques.

Finally, we believe that the novel extraction Keyphrases approach and the comparative study presented in this paper should be very useful to support the research in the field any Arabic Text Mining applications; the main contribution of this paper is four manifolds:

1. We must mention that our experiments show the improvements of the clustering quality and time when we use the extracted Keyphrases with our novel approach instead of the full-text representation of documents ,
2. The stemming affects negatively the final results, it makes the representation of the document smaller and the clustering faster,
3. Cosine Similarity, Jaccard and Pearson Correlation measures are quite similar for finding more coherent clusters for all documents clustering algorithms,
4. Ward technique is effective than other linkage techniques for producing more coherent clusters using the Agglomerative Hierarchical algorithm.

Finally, in the perspective we suggest to compare and evaluate our proposed method for Arabic Keyphrases extraction with the KP-Miner Keyphrases extraction system.

## AUTHORS

**Issam SAHMOUDI** received his Computer Engineering degree from ECOLE NATIONALE DES SCIENCES APPLIQUÉES, FEZ (ENSA-FEZ). Now he is PhD Student in Laboratory of Information Science and Systems LSIS, at (E.N.S.A), Sidi Mohamed Ben Abdellah University (USMBA), Fez, Morocco. His current research interests include Arabic Text Mining Applications: Arabic Web Document Clustering and Browsing, Arabic Information and Retrieval Systems, and Arabic web Mining.

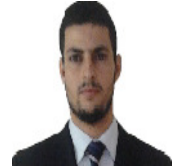

**Miss. Hanane Froud** Phd Student in Laboratory of Information Science and Systems, ECOLE NATIONALE DES SCIENCES APPLIQUÉES (ENSA-FEZ),University Sidi Mohamed Ben Abdellah (USMBA),Fez, Morocco.  She has also presented different papers at different National and International conferences.

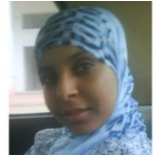

**Pr. Abdelmonaime LACHKAR** received his PhD degree from the USMBA, Morocco in 2004, He is  Professor and Computer Engineering Program Coordinator at (E.N.S.A, FES), and the Head of the Systems Architecture and Multimedia Team (LSIS Laboratory) at Sidi Mohamed Ben Abdellah University, Fez, Morocco. His current research interests include Arabic Natural Language Processing ANLP, Arabic Web Document Clustering and Categorization, Arabic Information Retrieval Systems, Arabic Text Summarization, Arabic Ontologies development and usage, Arabic Semantic Search Engines (SSEs).

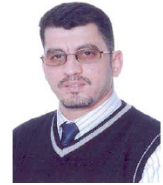